\documentclass[11pt,a4paper]{article}

\usepackage[utf8]{inputenc}
\usepackage[T1]{fontenc}

\usepackage{amsmath,amssymb}

\usepackage{graphicx}
\usepackage{booktabs}
\usepackage{url}
\usepackage{subcaption}
\usepackage[numbers,sort&compress]{natbib}
\usepackage{hyperref}
\hypersetup{hidelinks}

\title{Inducing Spatial Locality in Vision Transformers through the Training Protocol}

\author{Eduardo Antonio Santiago Toledo\\
	\small Universidad Autónoma Metropolitana\\
	\small \texttt{e.santiago@xanum.uam.mx}\\[4pt]
Asael Fabian Martínez\\
	\small Universidad Autónoma Metropolitana\\
	\small \texttt{fabian@xanum.uam.mx}}

\date{}

\begin{document}
	
	\maketitle
	
	% ==================== ABSTRACT ====================
	
	\begin{abstract}
		We investigate whether the training protocol can induce spatial locality in the early layers of a Vision Transformer (ViT) trained from scratch, without large-scale pretraining. Keeping the architecture and optimization procedure fixed, we compare a Baseline protocol with a Modern protocol (AutoAugment/ColorJitter, CutMix, and Label Smoothing) on CIFAR-10, CIFAR-100, and Tiny-ImageNet, characterizing each attention head via Mean Attention Distance (MAD) and normalized entropy. Across all three datasets, the Modern protocol produces more local and more concentrated attention in early layers; on CIFAR-100, the minimum MAD drops from 0.316 (Baseline) to 0.008 (Modern). To identify the source of this effect, we conduct an ablation study on CIFAR-100 by adding or removing each component individually. The results identify CutMix as the determining component within our experiments: all conditions with CutMix exhibit MAD $\leq 0.024$, while all conditions without CutMix remain at MAD $\geq 0.210$. AutoAugment and Label Smoothing show no independent effect on locality. Taken together, these findings suggest that the pressure to classify from partial image regions, induced by CutMix, can promote the emergence of local attention in Vision Transformers.
	\end{abstract}

	
	% ==================== 1. INTRODUCTION ====================
	\section{Introduction}
	\label{sec:intro}
	
	Vision Transformers (ViTs)~\cite{dosovitskiy2020image} have achieved competitive performance against convolutional neural networks (CNNs) in image classification. However, unlike CNNs, which explicitly incorporate a locality inductive bias through convolutions and limited receptive fields~\cite{lecun1998gradient,he2016deep}, ViTs operate with global self-attention: every token can attend to every other token without spatial restriction. This flexibility can in principle capture long-range dependencies, but it also means the model must learn to attend locally if doing so is useful for the task. This raises a central question: \textit{under what conditions does local behavior emerge in the early layers of a ViT?}
	
	Raghu et al.~\cite{raghu2021dovit} addressed this question by comparing ViTs pretrained on JFT-300M (approximately 300 million images) against ViTs trained solely on ImageNet-1k (1.2 million images). Using the Mean Attention Distance (MAD) metric proposed by Dosovitskiy et al.~\cite{dosovitskiy2020image}, they showed that models pretrained on JFT-300M develop attention heads with local behavior in lower layers (that is, heads that concentrate their attention on spatially nearby tokens). However, the emergence of this pattern depends on the relationship between model capacity and data quantity: high-capacity models (ViT-L/16 and ViT-H/14) trained only on ImageNet-1k fail to learn local attention in early layers, while a smaller model (ViT-B/32) does develop some degree of locality with the same dataset. This suggests that locality does not depend exclusively on data scale in absolute terms and motivates the exploration of additional factors that may induce this behavior.
	
	In this work, we propose that the training protocol (specifically, the augmentation and regularization techniques applied during training) can serve as an alternative pathway for the emergence of locality, even on moderate-scale datasets. Our hypothesis is that certain modern training techniques can increase generalization pressure in a way that they incentivize early layers to extract local discriminative features, without the need for pretraining on millions of images.
	
	To evaluate this hypothesis, we design a controlled experiment in which a compact ViT is trained from scratch (without pretrained weights) under two protocols that share the same architecture and optimizer, differing only in augmentation and regularization techniques. The \textit{Baseline} protocol uses minimal augmentation (Random Crop and Random Horizontal Flip), while the \textit{Modern} protocol adds three components: AutoAugment~\cite{cubuk2019autoaugment} on CIFAR-10/100 or ColorJitter, CutMix~\cite{yun2019cutmix}, and Label Smoothing~\cite{muller2019label}. We evaluate both protocols on three datasets of increasing complexity: CIFAR-10, CIFAR-100, and Tiny-ImageNet (50{,}000--100{,}000 training images).
	
	To characterize the resulting attention patterns, we employ two complementary per-head metrics: Mean Attention Distance, which quantifies how far each head attends spatially on average, and normalized attention entropy, which measures how concentrated or diffuse each head's attention distribution is. Together, these metrics distinguish between heads that attend locally and in a concentrated manner (low MAD, low entropy) and heads that distribute their attention globally and uniformly (high MAD, high entropy).
	
	Our analysis proceeds in two stages. In the first (Section~\ref{sec:results}), we compare the Baseline and Modern protocols across all three datasets, establishing that the Modern protocol is consistently associated with lower MAD and lower entropy in early layers. The effect is particularly pronounced on CIFAR-100, where the minimum MAD drops from 0.316 (Baseline) to 0.008 (Modern). In the second stage (Section~\ref{sec:ablation}), we conduct an ablation study on CIFAR-100 to identify which of the three components of the Modern protocol is responsible for this effect. The results identify CutMix as the determining component for inducing locality: the separation between conditions with and without CutMix is complete (MAD $\leq 0.024$ vs.\ MAD $\geq 0.210$), while AutoAugment and Label Smoothing show no independent effect.
	
	We emphasize that we do not claim equivalence with large-scale pretraining; our results are limited to a compact architecture and moderate-scale datasets. Nevertheless, the identification of CutMix as the determining component within our experimental design provides a more precise understanding of how training design choices can shape the internal attention structure of Vision Transformers.
	
	The contributions of this work are as follows. First, we show empirically that a Modern training protocol is associated with the emergence of local attention in early layers of ViTs trained from scratch on 50{,}000--100{,}000 images, with consistent results across CIFAR-10, CIFAR-100, and Tiny-ImageNet. Second, we propose a per-head quantitative analysis combining MAD and normalized entropy, identifying a low-distance, low-entropy region in the Modern protocol that is absent in the Baseline protocol. Third, through an ablation study that evaluates each component in isolation, we identify CutMix as the only component that induces attentional locality, showing that neither AutoAugment nor Label Smoothing contribute independently to this effect.
	
	% ==================== 2. RELATED WORK ====================
	\section{Related Work}
	\label{sec:related}
	
	\subsection{Interpretability of Vision Transformers}
	The analysis of internal representations in Vision Transformers has been the subject of intensive research since their introduction. Raghu et al.~\cite{raghu2021dovit} conducted a comprehensive comparative analysis between ViTs and CNNs using Centered Kernel Alignment (CKA)~\cite{kornblith2019similarity}, revealing fundamental structural differences: ViTs exhibit greater uniformity in their representations across layers, while ResNets exhibit a clear hierarchical organization with low similarity between early and late layers.
	
	Using the attention distance metric proposed by Dosovitskiy et al.~\cite{dosovitskiy2020image}, Raghu et al~\cite{raghu2021dovit}. showed that models pretrained on JFT-300M (approximately 300 million images) develop attention heads with local behavior in lower layers, reflected in low average distances indicating restricted spatial focus. However, the emergence of this pattern depends on the relationship between model capacity and available data. When high-capacity models (ViT-L/16 and ViT-H/14) are trained solely on ImageNet-1k (1.2 million images), their early layers fail to learn local attention, exhibiting uniformly high distances. In contrast, ViT-B/32, a smaller model with coarser patches, does develop some degree of locality even with ImageNet alone~\cite{raghu2021dovit}. This suggests that the emergence of locality does not depend exclusively on data scale in absolute terms, but rather on the data being sufficient for the model's capacity, and motivates the exploration of additional factors that may induce this behavior.
	
	Our work complements these analyses by focusing specifically on attention mechanism metrics, namely Mean Attention Distance and attention entropy, to characterize the local versus global behavior of attention heads. Unlike CKA, which compares full-layer representations, MAD provides an interpretable and granular measure of the emergent inductive bias at the level of individual attention heads. We adopt the methodology of Raghu et al. but apply it in a different context: instead of varying data scale, we vary the training protocol and, through an ablation study, isolate the component responsible for the observed effect.
	
	\subsection{Modern Training Protocols}
	
	Effective training of Vision Transformers has undergone rapid evolution since the seminal work of Dosovitskiy et al.~\cite{dosovitskiy2020image}, who showed that, under their original configuration, the competitive performance of ViTs depended heavily on large-scale pretraining (JFT-300M or ImageNet-21k). This dependence on massive data motivated research into training strategies that could reduce this data efficiency gap.
	
	Touvron et al. introduced Data-efficient image Transformers (DeiT)~\cite{touvron2021deit}, demonstrating that Vision Transformers can be trained effectively on ImageNet-1k without external data. The authors note that, since Transformers lack the inductive biases inherent to CNNs, they typically require larger data volumes; to close this gap, extensive augmentation and strong regularization are essential~\cite{touvron2021deit}. DeiT validated that techniques such as automatic augmentation, image mixing (CutMix), stochastic regularization, and decoupled optimization can significantly reduce the need for large-scale pretraining, achieving substantial accuracy gains without external data.
	
	The main components of these modern protocols are as follows. AutoAugment~\cite{cubuk2019autoaugment} uses reinforcement learning to discover augmentation policies optimized per dataset, increasing effective diversity through geometric and color transformations. CutMix~\cite{yun2019cutmix} combines pairs of images by cutting and pasting rectangular regions, improving not only accuracy but also spatial localization ability. Label Smoothing~\cite{muller2019label} softens one-hot labels by combining them with a uniform distribution over all classes, which reduces the model's absolute confidence, improves calibration, and reduces overfitting when combined with aggressive augmentation.
	
	Additionally, AdamW~\cite{loshchilov2019adamw} and the warmup + cosine decay schedule~\cite{loshchilov2017sgdr,touvron2021deit} have become standard practice in Transformer training; in this work we keep them fixed across both protocols to isolate the effect of augmentation-based regularization (AutoAugment/ColorJitter), image mixing (CutMix), and label smoothing.
	
	These techniques have been widely adopted for training Vision Transformers without large-scale pretraining. However, their effect on the internal structure of attentional representations (e.g., the emergence of locality as measured by MAD) remains insufficiently characterized.
	
	\subsection{The Gap in the Literature}
	Despite the substantial accuracy improvements demonstrated by modern training techniques~\cite{touvron2021deit}, a systematic analysis of their effects on the internal structure of attentional representations in Vision Transformers is lacking. In particular, we identify two open questions.
	
	The first is whether the training protocol can be associated with the emergence of locality on moderate-scale datasets. Raghu et al.~\cite{raghu2021dovit} established that data scale is critical for this phenomenon, but their experiments kept the training protocol fixed and varied only the dataset size. Our work addresses this question through a controlled comparison between Baseline and Modern protocols on three datasets of 50{,}000--100{,}000 images.
	
	The second question is which component of the Modern protocol is responsible for the effect, if any. Augmentation and regularization techniques (AutoAugment, CutMix, Label Smoothing) are typically applied jointly, making it difficult to attribute effects to individual components. Our ablation study addresses this question by evaluating each component in isolation, allowing us to assess whether any single component is responsible for the emergence of locality.
	
	% ==================== 3. MATHEMATICAL FRAMEWORK ====================
	\section{Mathematical Framework} 
	\label{sec:math}
	
	To characterize the structure of the emergent inductive bias, we formalize two complementary per-head metrics: Mean Attention Distance (MAD), grounded in receptive field analysis~\cite{dosovitskiy2020image}, and normalized attention entropy, based on Shannon's information theory~\cite{shannon1948mathematical}. The former quantifies where each head attends in terms of spatial distance, and the latter measures how selective that attention is. In this section we formalize both metrics from the attention distribution of the multi-head mechanism.
	
	\subsection{Multi-Head Attention Mechanism}
	
	Following~\cite{vaswani2017attention}, let $\mathbf{X}\in\mathbb{R}^{N\times D}$ be the input sequence with $N=N_p+1$ tokens (patches + \texttt{[CLS]}), where $N_p$ is the number of patches and $D$ the embedding dimension. Each head $h$ in layer $l$ projects the input into queries and keys via learnable matrices $\mathbf{W}_Q^{(h,l)}, \mathbf{W}_K^{(h,l)} \in \mathbb{R}^{D \times d_k}$, where $d_k = D/H$ is the per-head dimension and $H$ the total number of heads. The query and key matrices are computed as $Q^{(h,l)} = \mathbf{X}^{(l)}\mathbf{W}_Q^{(h,l)}$ and $K^{(h,l)} = \mathbf{X}^{(l)}\mathbf{W}_K^{(h,l)}$, respectively. The score matrix $\mathbf{S}^{(h,l)}$ and the attention distribution $\mathbf{A}^{(h,l)}$ are then defined as:
	\begin{equation}
		\mathbf{S}^{(h,l)}=\frac{Q^{(h,l)}(K^{(h,l)})^\top}{\sqrt{d_k}}, 
		\quad 
		\mathbf{A}^{(h,l)}=\operatorname{softmax}\!\left(\mathbf{S}^{(h,l)}\right),
	\end{equation}
	where $\operatorname{softmax}$ is applied row-wise for $i=1,\dots,N$, so that $\sum_{j=1}^{N}A^{(h,l)}_{ij}=1$. The factor $1/\sqrt{d_k}$ stabilizes gradients by scaling the dot products by the square root of the head dimension.
	
	The resulting matrix $\mathbf{A}^{(h,l)} \in \mathbb{R}^{N \times N}$ encodes the relative importance that each token assigns to all others. Our analysis focuses on this attention distribution because it allows direct examination of the spatial connectivity pattern that each head has learned.
	
	The \texttt{[CLS]} token is a classification token with no defined spatial position, whose role is to aggregate global information for the final prediction. Since our analysis focuses on the spatial relationships between patches (that is, whether attention concentrates on nearby neighbors or is distributed globally), including \texttt{[CLS]} would distort the spatial metrics by introducing a token with no grid coordinate. We therefore work exclusively with patch-to-patch interactions.
	
	To establish notation, we assign index $1$ to the \texttt{[CLS]} token and indices $2,\dots,N$ to the patches. We define the renormalized attention matrix $\widetilde{\mathbf{A}}^{(h,l)} \in \mathbb{R}^{N_p \times N_p}$ excluding \texttt{[CLS]}. For $i,j\in\{1,\dots,N_p\}$,
	\begin{equation}
		\widetilde{A}_{ij}^{(h,l)} = \frac{A_{i+1, j+1}^{(h,l)}}{\sum_{k=1}^{N_p} A_{i+1, k+1}^{(h,l)}}, 
		\quad \text{such that} \quad \sum_{j=1}^{N_p}\widetilde{A}_{ij}^{(h,l)}=1.
	\end{equation}
	Renormalization is necessary because, upon excluding the column corresponding to \texttt{[CLS]}, the probability mass originally assigned to that token is redistributed proportionally among the remaining patches, preserving the interpretation of $\widetilde{A}_{ij}^{(h,l)}$ as a valid probability distribution over spatial positions.
	
	\subsection{Mean Attention Distance}
	
	Mean Attention Distance quantifies the average spatial distance at which each head attends, weighted by the attention weights~\cite{dosovitskiy2020image,raghu2021dovit}. Let $\mathbf{p}_i \in \mathbb{R}^2$ be the spatial coordinate of patch $i$ on a $G \times G$ grid with $G=\sqrt{N_p}$. Coordinates are assigned according to the row-column position of the patch in the grid (for example, on an $8\times 8$ grid, the top-left patch has $\mathbf{p}_1 = (0, 0)$ and the bottom-right patch has $\mathbf{p}_{64} = (7, 7)$). We define the normalized Euclidean distance $\delta(i, j)$ as:
	\begin{equation}
		\delta(i, j) = \frac{\|\mathbf{p}_i - \mathbf{p}_j\|_2}{\Delta_{\text{grid}}}, 
		\qquad 
		\Delta_{\text{grid}}=\sqrt{2}(G-1).
	\end{equation}
	The denominator $\Delta_{\text{grid}}$ corresponds to the maximum possible Euclidean distance on the grid (the diagonal between opposite corners, e.g., $(0,0)$ and $(G-1,G-1)$), which guarantees $\delta(i, j)\in[0,1]$. This normalization allows direct comparability across datasets with the same patch geometry (in our case, $8\times 8$ for all three datasets). The MAD for head $h$ in layer $l$ is:
	\begin{equation}
		\text{MAD}(h, l) = \frac{1}{N_p} \sum_{i=1}^{N_p} \sum_{j=1}^{N_p} \widetilde{A}_{ij}^{(h,l)} \cdot \delta(i, j).
	\end{equation}
	
	Intuitively, MAD averages over all query tokens $i$ the expected distance to the attended token, weighted by attention. The extreme cases illustrate its interpretation: if a head concentrates all attention from each token onto itself ($\widetilde{A}_{ij} = \mathbf{1}_{[i=j]}$), then $\text{MAD} = 0$ (perfectly local attention). If attention were distributed so as to maximize distance, MAD would approach its geometric maximum, which is bounded above by~$1$. A head with uniform attention ($\widetilde{A}_{ij} = 1/N_p$) produces an intermediate value reflecting the average distance between all pairs of patches on the grid.
	
	\subsection{Attention Entropy}
	
	While MAD measures the spatial extent of attention, entropy quantifies its selectivity: a head may attend locally but in a diffuse manner (distributed across many nearby neighbors) or in a concentrated manner (focused on a few specific tokens). To capture this distinction, we employ Shannon entropy~\cite{shannon1948mathematical} applied to the renormalized attention distribution.
	
	From $\widetilde{\mathbf{A}}^{(h,l)}\in\mathbb{R}^{N_p\times N_p}$, we interpret each row $i$ as a probability distribution over the patches $j$ attended by query patch $i$, i.e., $\sum_{j=1}^{N_p}\widetilde{A}_{ij}^{(h,l)}=1$. We define the Shannon entropy per query token $i$ as:
	\begin{equation}
		H_i^{(h,l)}=-\sum_{j=1}^{N_p}\widetilde{A}_{ij}^{(h,l)}\,
		\ln\!\big(\widetilde{A}_{ij}^{(h,l)}+\epsilon\big),
	\end{equation}
	where $\epsilon=10^{-12}$ is introduced to avoid numerical instability when $\widetilde{A}_{ij}^{(h,l)}\approx 0$.
	
	Finally, the attention entropy of head $(h,l)$ is obtained by averaging $H_i^{(h,l)}$ over all query tokens and normalizing by $\ln N_p$, so that the metric lies approximately in $[0,1]$, up to a negligible numerical error due to the $\epsilon=10^{-12}$ stabilization, that is:
	\begin{equation}
		\text{Entropy}(h,l)=\frac{1}{\ln N_p}\cdot \frac{1}{N_p}\sum_{i=1}^{N_p} H_i^{(h,l)}.
	\end{equation}
	Here, the normalization by $\ln N_p$ (the maximum entropy of a uniform distribution over $N_p$ elements) ensures that the metric is interpretable regardless of the number of patches: $\text{Entropy}(h,l) \approx 0$ indicates an attention distribution concentrated on a few tokens (in the extreme, a single token receives all attention), while $\text{Entropy}(h,l) \approx 1$ indicates attention uniformly distributed over all patches.
	
	\subsection{Complementarity of the Metrics}
	
	MAD and entropy capture distinct aspects of attentional behavior, and their joint use provides a more complete characterization than either metric alone. MAD only indicates the spatial extent of attention, not its selectivity: entropy distinguishes whether a head with low MAD concentrates its attention on a single neighbor (low entropy, analogous to a convolutional filter with a small receptive field) or distributes it across many nearby neighbors (high entropy). Symmetrically, MAD complements entropy: a head with low entropy could focus its attention on a distant token or on an immediate neighbor, and only MAD discriminates between these two scenarios. In Section~\ref{sec:results}, the two-dimensional MAD--entropy space reveals patterns that neither metric would identify in isolation.
	
	Finally, we note that Equations (4) and (6) define MAD and entropy for a single image. In practice, we average both metrics over a set of $n$ evaluation images to obtain stable estimates (see Section~\ref{sec:exp}).
	
	% ==================== 4. EXPERIMENTAL DESIGN ====================
	\section{Experimental Design}
	\label{sec:exp}
	
	To evaluate whether a training protocol can be associated with the emergence of locality in Vision Transformers without relying on massive datasets, we design a controlled experiment comparing two protocols (Baseline vs.\ Modern) on three datasets of increasing complexity (CIFAR-10, CIFAR-100, Tiny-ImageNet). In all cases we train from scratch (without pretrained weights), controlling architecture and optimization to attribute any observed differences exclusively to the regularization and augmentation protocol.
	
	\subsection{Model Architecture}
	
	We employ a compact Vision Transformer inspired by~\cite{dosovitskiy2020image}. Given a batch of images, the model:
	(i) divides the image into non-overlapping patches via a \texttt{Conv2d} projection with \texttt{kernel\_size} = \texttt{stride} = \texttt{patch\_size};
	(ii) flattens the patch grid into a sequence of $N_p$ tokens; 
	(iii) prepends a learnable \texttt{[CLS]} token and adds learnable positional embeddings; 
	(iv) processes the sequence through $L$ Pre-LN Transformer blocks; and 
	(v) classifies using exclusively the final embedding of the \texttt{[CLS]} token.
	
	The architectural hyperparameters are fixed as follows: depth $L = 8$ Transformer blocks, embedding dimension $D = 192$, multi-head attention with $H = 8$ heads ($d_k = D/H = 24$ per head), internal MLP with hidden dimension $2D = 384$ and GELU activation, and internal dropout of $0.0$ in both attention and MLP.
	
	The choice of a compact architecture ($\approx$2.4M parameters) reflects two considerations. First, it is appropriate for the scale of the datasets used (50{,}000--100{,}000 images), where a larger model could overfit more easily. Second, it allows training all configurations in the study (two protocols on three datasets, plus six ablations) within a reasonable computational budget. Setting internal dropout to $0.0$ is a deliberate choice: by eliminating all stochastic regularization within the network, any observed effect on attention patterns can be attributed exclusively to differences in the training protocol (data augmentation and label smoothing), without confounding dropout effects.
	
	\subsection{Datasets and Preprocessing}
	
	We select three datasets covering an increasing range of complexity in number of classes and resolution: CIFAR-10~\cite{krizhevsky2009learning} (10 classes, 50{,}000 training images and 10{,}000 test images, $32\times 32$ resolution), CIFAR-100~\cite{krizhevsky2009learning} (100 classes, 50{,}000 training and 10{,}000 test, $32\times 32$ resolution), and Tiny-ImageNet~\cite{le2015tinyimagenet} (200 classes, 100{,}000 training and 10{,}000 validation with no public labeled test set, $64\times 64$ resolution).
	
	This selection allows us to evaluate whether the effect of the training protocol is consistent across different levels of class granularity (10, 100, 200) and image resolution ($32\times 32$, $64\times 64$), while keeping data scale in the moderate range of 50{,}000 to 100{,}000 images.
	
	To prevent information leakage, we partition the official training set into a stratified 80/20 split: 80\% for training and 20\% for internal validation. Checkpoint selection is based exclusively on validation accuracy (consistent with the early stopping criterion), and the test set is used only for final evaluation. For Tiny-ImageNet, where no public labeled test set exists, the official validation split serves as the final test set. All partitioning and shuffling is controlled with a fixed seed to ensure reproducibility.
	
	To enable direct comparison of spatial metrics (MAD/entropy) across datasets, we fix the number of spatial tokens to $N_p=64$ in all cases: CIFAR-10/100 uses $4\times 4$ patches on $32\times 32$ images, and Tiny-ImageNet uses $8\times 8$ patches on $64\times 64$ images, both resulting in an $8\times 8$ grid with 64 patches. The total sequence length is thus $N=N_p+1=65$ including \texttt{[CLS]}, and the distance matrix $\delta(i,j)$ defined in Section~\ref{sec:math} is identical for all three datasets. Images are normalized using the standard statistics of each dataset.
	
	\subsection{Training Protocols}
	
	We define two training protocols for comparison. In both cases we keep the architecture, base optimizer, and learning-rate schedule fixed; the differences are limited to augmentation and loss regularization.
	
	\subsubsection{Baseline Protocol}
	This represents a conservative training setup without aggressive modern regularization. We use the AdamW optimizer~\cite{loshchilov2019adamw} with learning rate $\eta=0.002$ (CIFAR-10/100) and $\eta=0.001$ (Tiny-ImageNet, reduced for stability with the higher input resolution), weight decay $\lambda=5\times 10^{-5}$ (CIFAR) and $\lambda=1\times 10^{-4}$ (Tiny-ImageNet), and a linear warmup + cosine decay schedule~\cite{loshchilov2017sgdr,touvron2021deit} with 5 warmup epochs (CIFAR) and 10 epochs (Tiny-ImageNet), followed by cosine annealing to the maximum number of epochs. The loss is standard \texttt{CrossEntropyLoss} (without label smoothing) and augmentation is limited to Random Crop + Random Horizontal Flip~\cite{shorten2019survey}. The batch size is 128 (CIFAR) and 64 (Tiny-ImageNet), with a maximum of 600 epochs (CIFAR) and 400 (Tiny-ImageNet). Early stopping is applied by monitoring validation accuracy, with a patience of 20 epochs (CIFAR) and 30 (Tiny-ImageNet).
	
	\subsubsection{Modern Protocol}
	This incorporates regularization and augmentation that are standard in efficient Transformer training~\cite{touvron2021deit}. The optimizer, learning rates, weight decay, and schedule are identical to the Baseline protocol. On CIFAR-10/100, augmentation includes Random Crop + Random Horizontal Flip + AutoAugment~\cite{cubuk2019autoaugment} (\texttt{CIFAR10} policy applied consistently on both CIFAR datasets); on Tiny-ImageNet, AutoAugment is replaced by ColorJitter (realistic color and lighting variations for natural photography). CutMix~\cite{yun2019cutmix} is applied at the mini-batch level with $\alpha=1.0$ and $p=0.6$ on all datasets; in each application, $\lambda$ is sampled from a Beta($\alpha,\alpha$) distribution and the target label is linearly interpolated based on the actual area of the mixed patch. Label Smoothing~\cite{muller2019label} is incorporated with $\epsilon=0.05$. The batch size, maximum epochs, and early stopping are identical to the Baseline protocol.
	
	In summary, our comparison is controlled: observed changes in attention patterns (MAD/entropy) or accuracy reflect the joint effect of AutoAugment/ColorJitter, CutMix, and Label Smoothing, while the architecture and the rest of the optimization procedure remain fixed.
	
	On CIFAR-10/100 we use AutoAugment because a standard and widely used policy exists for $32\times 32$ images, enabling strong augmentation comparable to the literature. On Tiny-ImageNet we use ColorJitter because it is a simple and stable option for $64\times 64$ natural photography, introducing realistic color and lighting variations. In this way, we maintain strong augmentation in the Modern protocol while adapting it to the domain of each dataset.
	
	\subsection{Performance Validation}
	
	Before analyzing attention patterns, we verify that the Modern protocol improves generalization. We report Top-1 accuracy on the test set, evaluated using the checkpoint with the best validation accuracy (consistent with the early stopping criterion). Table~\ref{tab:model_performance} summarizes the results.
	
	\begin{table}[h]
		\centering
		\caption{Top-1 accuracy (\%) on test using the checkpoint with best validation. The Modern protocol consistently improves generalization across all three datasets.}
		\label{tab:model_performance}
		\begin{tabular}{lccc}
			\toprule
			\textbf{Dataset} & \textbf{Baseline (\%)} & \textbf{Modern (\%)} & \textbf{Gain ($\Delta$)} \\
			\midrule
			CIFAR-10      & 78.46 & 84.84 & +6.38 \\
			CIFAR-100     & 48.53 & 63.38 & +14.85 \\
			Tiny-ImageNet & 34.89 & 41.92 & +7.03 \\
			\bottomrule
		\end{tabular}
	\end{table}
	
	These gains (between +6.38 and +14.85 percentage points) confirm that the Modern protocol produces more robust representations. The most pronounced effect is observed on CIFAR-100, where the larger number of classes appears to amplify the benefit of regularization. In the following sections we analyze whether this improvement is associated with structural changes in attention, particularly the emergence of highly local heads in early layers.
	
	\subsection{Evaluation Procedure for MAD and Entropy}
	
	For each dataset, we sample $n=5{,}000$ images from the test set, identical across protocols. For Tiny-ImageNet, this subset is drawn from the official validation split that serves as the final test set. For each image, we extract the attention matrices $\mathbf{A}^{(h,l)}\in \mathbb{R}^{N\times N}$ in evaluation mode (without dropout or augmentation). We exclude \texttt{[CLS]} and renormalize according to Equation (2) to obtain $\widetilde{\mathbf{A}}^{(h,l)} \in \mathbb{R}^{64\times 64}$, and compute MAD and entropy according to Equations (4) and (6), respectively. The values reported for each head correspond to the average over the 5{,}000 images, computed in \texttt{float64} precision; the variability across images is reflected in the confidence bands of Figure~\ref{fig:mad_profiles}. All partitioning, shuffling, and sampling is controlled with a global seed. The entire implementation (architecture, training, evaluation, and metric computation) was carried out in PyTorch~\cite{paszke2019pytorch}.
	
	This experimental design allows any change in locality (MAD) and concentration (entropy) to be cleanly attributed to the training protocol, while keeping the architecture, the number of spatial tokens, and the evaluation procedure constant.
	
	% ==================== 5. RESULTS AND ANALYSIS ====================
	\section{Results and Analysis}
	\label{sec:results}
	
	In this section, we present a quantitative analysis of the learned attention patterns. We organize the evidence along three dimensions: (i) MAD profiles, which assess the consistency of locality across datasets; (ii) Attentional behavior space (MAD--Entropy), which characterizes the behavior of each attention head; and (iii) Heatmaps, which visualize the organization of locality and dispersion along depth.
	
	\subsection{Distance Profile Analysis}
\begin{figure}[t]
		\centering
		\begin{subfigure}[b]{0.48\textwidth}
			\centering
			\includegraphics[width=\textwidth]{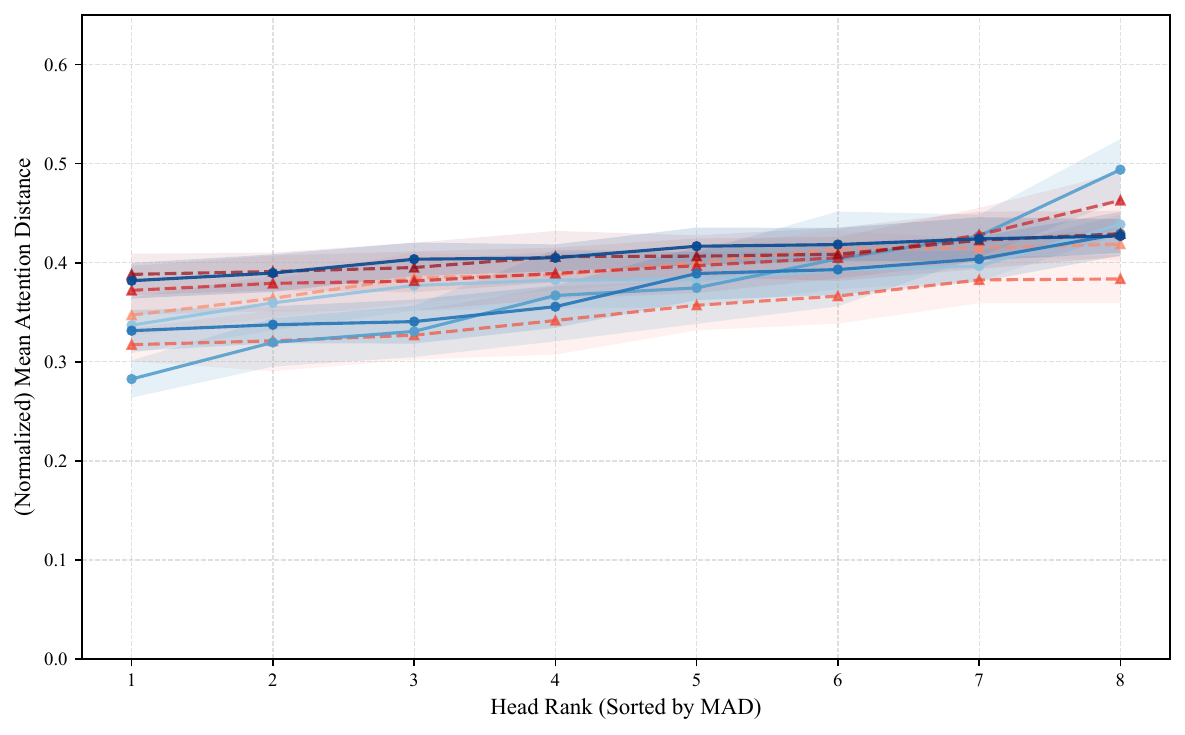}
			\caption{CIFAR-10}
		\end{subfigure}
		\hfill
		\begin{subfigure}[b]{0.48\textwidth}
			\centering
			\includegraphics[width=\textwidth]{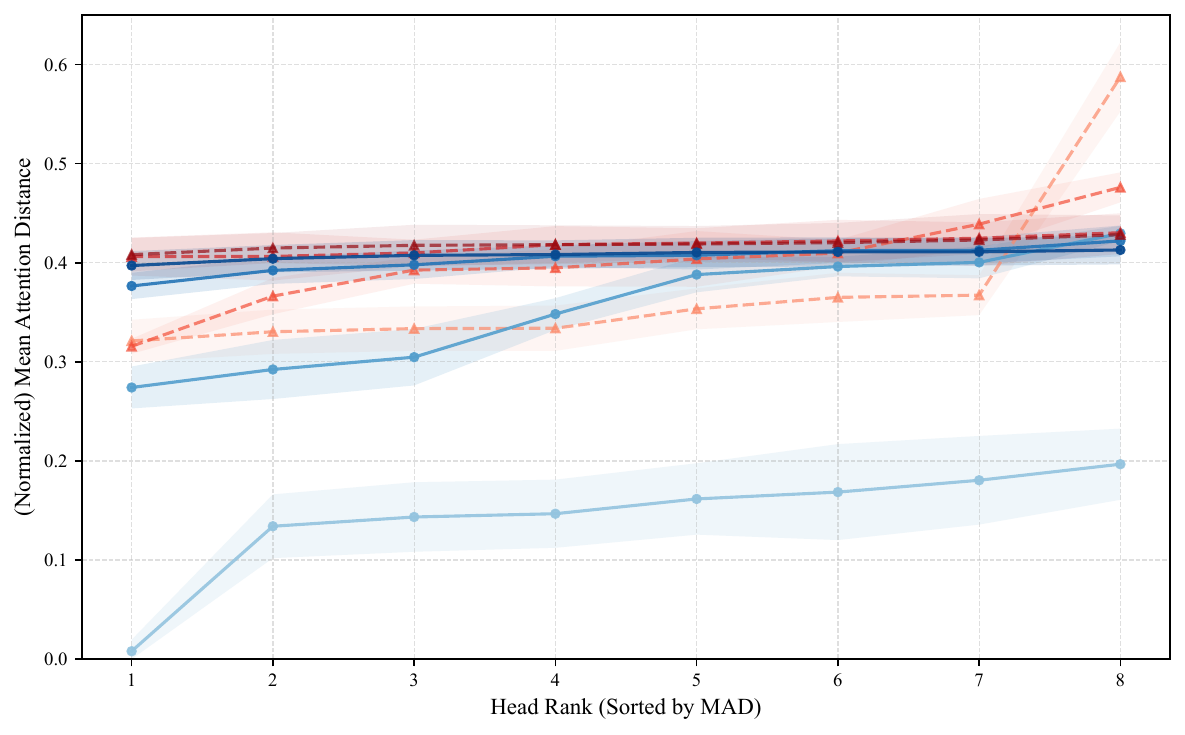}
			\caption{CIFAR-100}
		\end{subfigure}
		
		\vspace{0.5em}
		
		\begin{subfigure}[b]{0.48\textwidth}
			\centering
			\includegraphics[width=\textwidth]{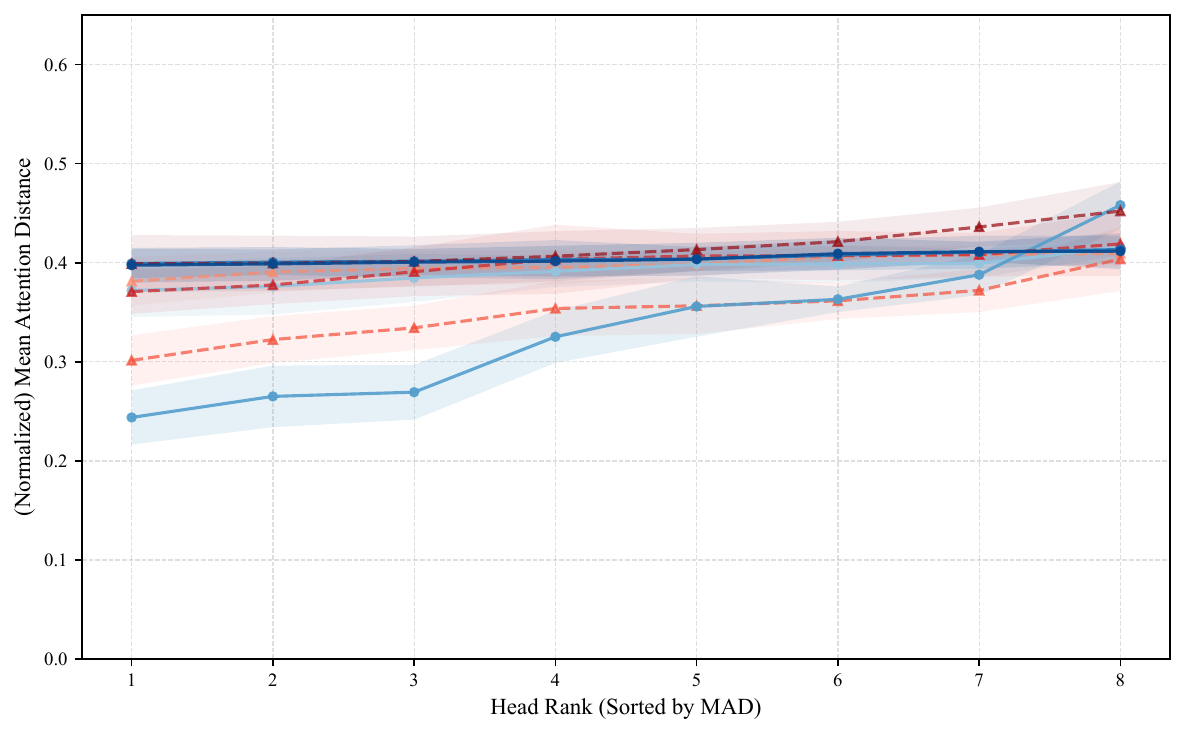}
			\caption{Tiny-ImageNet}
		\end{subfigure}
		\caption{MAD profiles for CIFAR-10, CIFAR-100, and Tiny-ImageNet. Each panel shows four layers (1, 2, 7, and 8), selected to contrast the extremes of the depth hierarchy. Within each layer, the eight heads are sorted from lowest to highest MAD (X-axis: Rank 1--8), resulting in eight lines per panel: four from the Modern protocol (blue, solid, circular markers) and four from the Baseline protocol (red, dashed, triangular markers). The shade lightens for early layers and darkens for deep layers. On CIFAR-100, Layer 1 of the Modern protocol (lightest blue line) drops to values near zero, while all Baseline lines remain above $\approx 0.31$.} 
		\label{fig:mad_profiles}
	\end{figure}

	We first analyze whether locality appears consistently across datasets of varying complexity. Figure~\ref{fig:mad_profiles} presents MAD values for four representative layers, the two earliest (L1, L2) and the two deepest (L7, L8), selected to contrast the extremes of the depth hierarchy. Within each layer, the eight heads are sorted from lowest to highest MAD, so that each line shows the sorted locality profile of a layer under a given protocol. Each panel therefore contains eight lines: four from the Modern protocol (blue, solid, circular markers) and four from the Baseline protocol (red, dashed, triangular markers); the shade lightens for early layers and darkens for deep layers.
	
	On CIFAR-100 (Figure~\ref{fig:mad_profiles}b), we observe the most pronounced effect. Under the Modern protocol, Layer 1 (lightest blue line) contains the head with the lowest MAD in the entire model ($\approx 0.008$), indicating attention focused almost exclusively on the patch itself or its immediate neighbors. In contrast, the Baseline lines (red, dashed) remain uniformly above MAD $\approx 0.31$, with no layer developing local attention. The relative difference between the global minima of the two protocols (0.008 vs.\ 0.316) is $97.5\%$.
	
	For CIFAR-10 and Tiny-ImageNet, while the effect is less extreme, the trend holds: the Modern protocol systematically induces greater locality. On CIFAR-10, the minimum MAD drops from $0.317$ (Baseline) to $0.199$ (Modern). Similarly, on Tiny-ImageNet the reduction is from $0.301$ to $0.244$. These results confirm that the Modern protocol consistently restricts the receptive field of early layers, although the magnitude of the effect varies across datasets and is most pronounced on CIFAR-100.
	
	\subsection{Attentional Behavior Space}
	\label{sec:scatter}
	
\begin{figure}[t]
		\centering
		\includegraphics[width=0.8\linewidth]{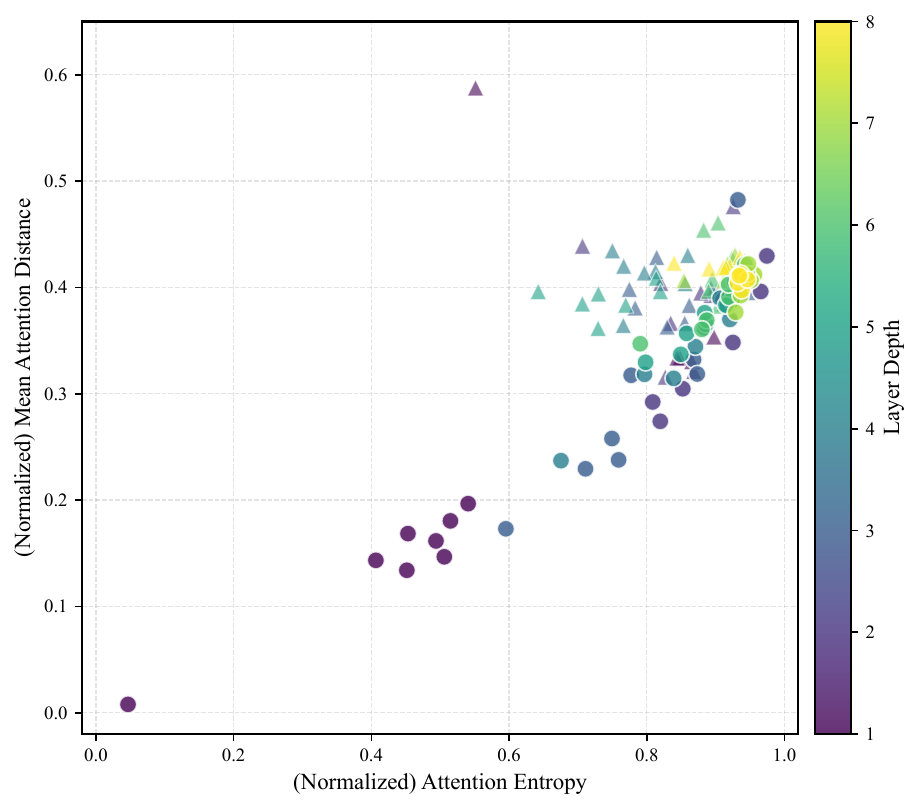}
		\caption{Attentional behavior space (CIFAR-100). Each point represents an attention head. Triangles correspond to the Baseline protocol and circles to the Modern protocol; color indicates layer depth (purple: L1 $\to$ yellow: L8). The Baseline concentrates in the upper-right region, indicating global and diffuse attention across all layers. The Modern protocol shows a structured trajectory: early layers descend toward the lower-left corner, reaching extreme levels of locality and concentration (MAD $\approx 0.008$, Entropy $\approx 0.047$) unattainable by the Baseline.}
		\label{fig:scatter}
	\end{figure}

	To understand the nature of this locality, we analyze the joint distribution of entropy and MAD. Figure~\ref{fig:scatter} visualizes each attention head of CIFAR-100 as a point in this two-dimensional space.
	
	The visualization reveals two completely distinct regimes. The Baseline protocol (triangles) shows a compact distribution: heads from all layers, from the first to the last, cluster in the high-entropy ($>0.55$) and high-distance ($>0.31$) region. This indicates that, without the pressure of modern regularization, the model operates globally and diffusely from the start, without developing a clear spatial hierarchy.
	
	In contrast, the Modern protocol (circles) displays a trajectory of progressive specialization. Early layers (purple and dark blue tones) separate sharply from the main cluster, shifting toward the lower-left corner. In particular, one head from Layer 1 (isolated purple circle) reaches values close to the origin (MAD $\approx 0.008$, Entropy $\approx 0.047$), indicating that it concentrates nearly all its attention on a few immediately adjacent patches.
	
	As depth increases (green to yellow tones), the circles ascend along the diagonal and eventually converge with the Baseline cluster. This behavior suggests that the Modern protocol does not eliminate the global capacity of the Transformer, but rather induces an initial stage of local processing that integrates progressively, functionally replicating the hierarchical structure (local $\to$ global) typical of convolutional networks.
	
	\subsection{Heatmaps}
	\label{sec:heatmaps}
	
\begin{figure}[t]
		\centering
		
		% --- Row 1: MAD ---
		\begin{subfigure}[b]{0.48\textwidth}
			\centering
			\includegraphics[width=\textwidth]{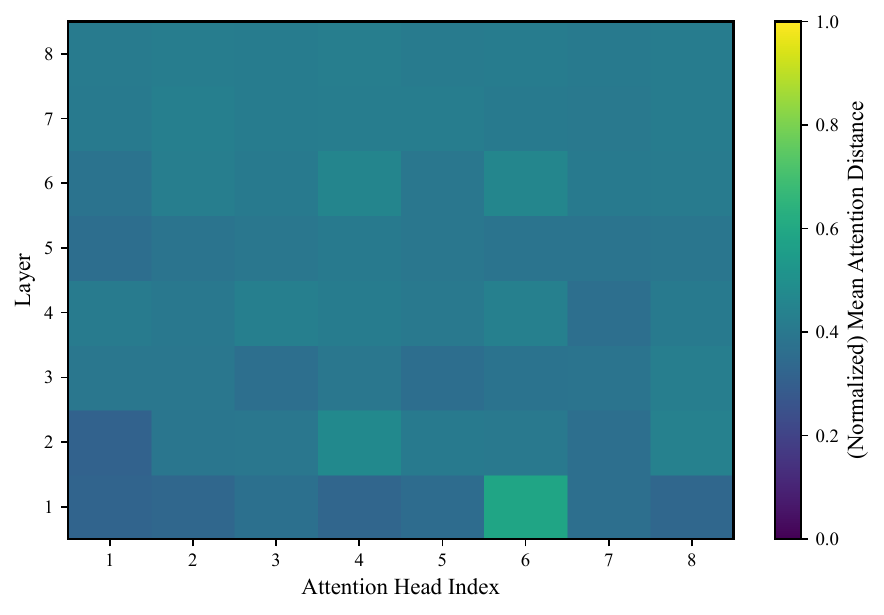}
			\caption{MAD (Baseline)}
		\end{subfigure}
		\hfill
		\begin{subfigure}[b]{0.48\textwidth}
			\centering
			\includegraphics[width=\textwidth]{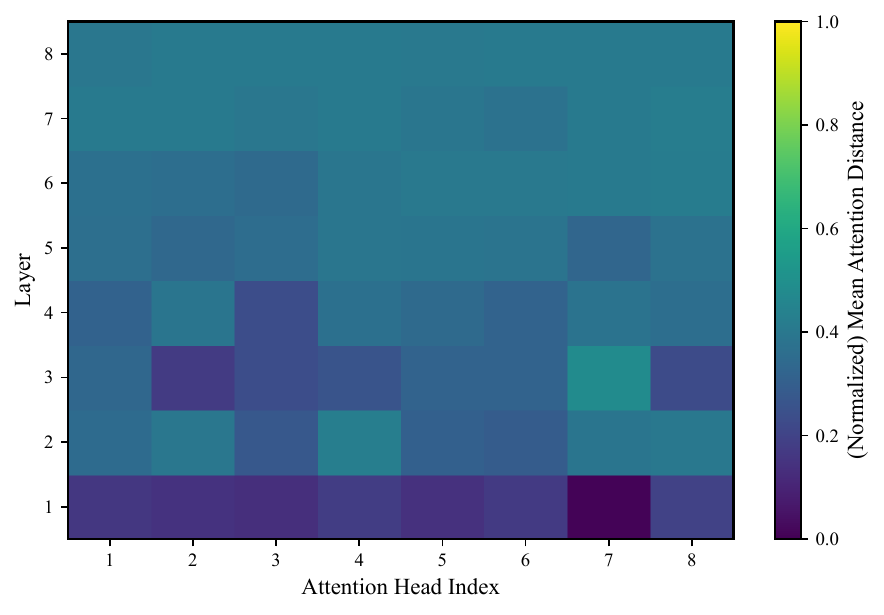}
			\caption{MAD (Modern)}
		\end{subfigure}
		
		\vspace{0.5em}
		
		% --- Row 2: Entropy ---
		\begin{subfigure}[b]{0.48\textwidth}
			\centering
			\includegraphics[width=\textwidth]{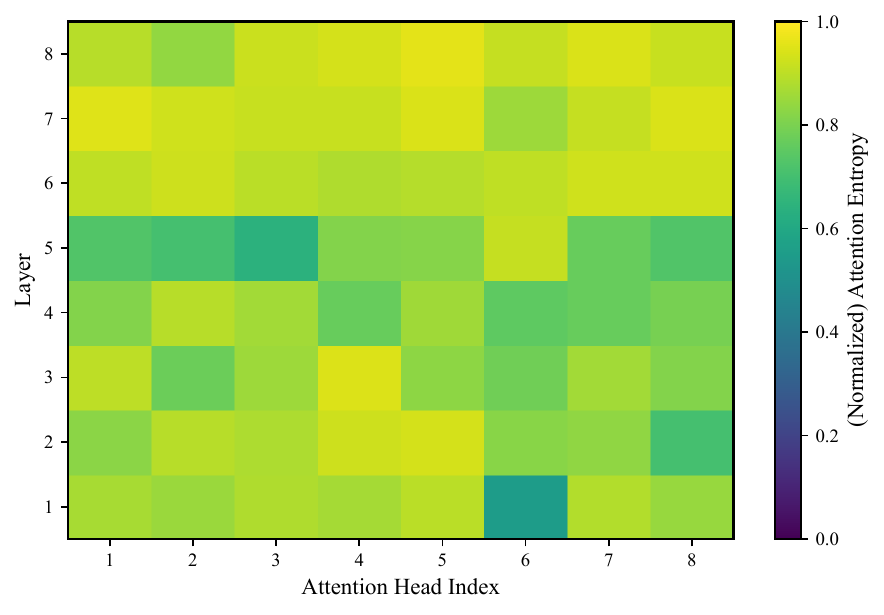}
			\caption{Entropy (Baseline)}
		\end{subfigure}
		\hfill
		\begin{subfigure}[b]{0.48\textwidth}
			\centering
			\includegraphics[width=\textwidth]{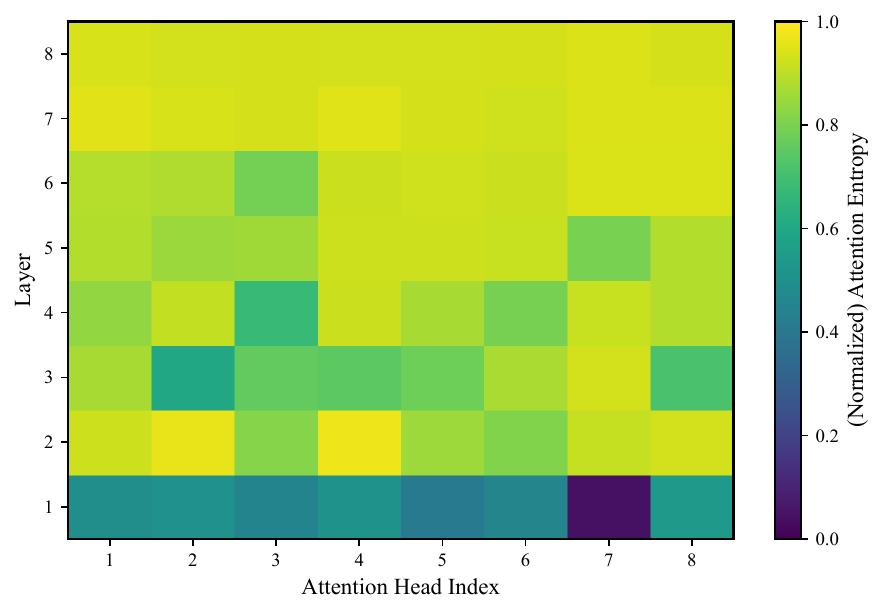}
			\caption{Entropy (Modern)}
		\end{subfigure}
		
		\caption{Heatmaps for CIFAR-100 (Y-axis: Layers 1--8; X-axis: Heads 1--8). The left column corresponds to the Baseline protocol and the right column to the Modern protocol. In the Baseline, moderate variation is observed without a clear hierarchical pattern across layers, and both distance (MAD) and concentration (Entropy) are relatively homogeneous. In the Modern protocol, a clear hierarchical structure emerges: Layer 1 (bottom row) shows a dark band of extreme locality in the MAD map and several heads with notably lower entropy in the entropy map; as depth increases, attention gradually becomes more global and diffuse.}
		\label{fig:heatmaps}
	\end{figure}

	Finally, Figure~\ref{fig:heatmaps} compares the internal organization of all heads via heatmaps, providing a complete view of the $8 \times 8 = 64$ attention heads in the model.
	
	The Baseline protocol (left column) shows a notable absence of hierarchical structure. The MAD and entropy maps exhibit moderate variation, but without a clear pattern between early and deep layers, maintaining intermediate values (teal/green colors) throughout the network.
	
	The Modern protocol (right column), by contrast, exhibits a clear organization. In the MAD map (Fig.~\ref{fig:heatmaps}b), Layer 1 appears as a dark purple band, confirming that all heads in the first layer have learned to restrict their spatial attention.
	
	The vertical gradient is equally informative: from Layer 1 to Layer 8, there is a general trend toward lighter colors with increasing depth. This indicates that the Modern protocol not only induces locality but produces a depth hierarchy in which the effective receptive field expands layer by layer.
	
	The entropy map (Fig.~\ref{fig:heatmaps}d) corroborates this observation with an additional detail: in Layer 1, head 7 shows a much darker cell, corresponding to an entropy close to zero. This indicates a highly concentrated attention mechanism at the input, contrasting with the uniform and diffuse distribution observed in the Baseline protocol.
	
	\subsection{Synthesis}
	
	The three analyses converge on a consistent conclusion. The MAD profiles establish that the Modern protocol reduces attention distance in early layers across all three datasets, with the most pronounced effect on CIFAR-100. The two-dimensional MAD--entropy space reveals that this reduction is accompanied by greater concentration of attention, forming a low-distance, low-entropy region exclusive to the Modern protocol. The heatmaps confirm that this pattern is organized as a depth hierarchy, with maximum locality in Layer 1 that relaxes progressively toward upper layers.
	
	These results demonstrate that the Modern protocol induces locality in early layers, but do not identify which of its three components (AutoAugment/ColorJitter, CutMix, Label Smoothing) is responsible for the effect. Section~\ref{sec:ablation} addresses this question through a systematic ablation study.
	
	% ==================== 6. ABLATION STUDY ====================
	\section{Ablation Study}
	\label{sec:ablation}
	
	To determine which of the three components of the Modern protocol (AutoAugment, CutMix, or Label Smoothing) is responsible for the emergence of locality, we conduct a systematic ablation study. We choose CIFAR-100 as the reference dataset because it is where the difference between protocols was most pronounced (minimum MAD of 0.008 vs.\ 0.316), which maximizes the available signal for discriminating among components.
	
	\subsection{Study Design}
	
	The study comprises six ablations organized in two complementary groups. In the first, we start from the Modern protocol and remove one component at a time while keeping the other two: $-$CutMix (AutoAugment + Label Smoothing), $-$AutoAugment (CutMix + Label Smoothing), and $-$Label Smoothing (CutMix + AutoAugment). In the second, we start from the Baseline protocol and add a single component: $+$CutMix, $+$AutoAugment, or $+$Label Smoothing.
	
	Together with the Baseline (no components) and Modern (all three components) protocols, these six ablations complete all eight possible presence/absence combinations of the three components, allowing evaluation of both the individual effect of each technique and their possible interactions.
	
	All other hyperparameters remain identical to those described in Section~\ref{sec:exp}. MAD and entropy metrics are computed on the same subset of 5{,}000 test images with a fixed seed.
	
	\subsection{Results}
	
	Table~\ref{tab:ablation} presents the complete results of the ablation study, including the Baseline and Modern protocols as references.
	
	\begin{table}[h]
		\centering
		\caption{Ablation study on CIFAR-100. Test accuracy and the range of MAD and entropy (minimum--maximum) across all attention heads are reported.}
		\label{tab:ablation}
		\begin{tabular}{lccc}
			\toprule
			\textbf{Condition} & \textbf{Test Acc (\%)} & \textbf{MAD} & \textbf{Entropy} \\
			\midrule
			Baseline (reference) & 48.53 & 0.316 -- 0.588 & 0.551 -- 0.958 \\
			Modern (reference) & 63.38 & 0.008 -- 0.482 & 0.047 -- 0.975 \\
			\midrule
			$-$ CutMix & 55.60 & 0.210 -- 0.449 & 0.644 -- 0.967 \\
			$-$ AutoAugment & 55.11 & 0.024 -- 0.492 & 0.106 -- 0.953 \\
			$-$ Label Smoothing & 60.34 & 0.009 -- 0.481 & 0.054 -- 0.959 \\
			\midrule
			$+$ CutMix & 55.12 & 0.021 -- 0.427 & 0.096 -- 0.941 \\
			$+$ AutoAugment & 55.22 & 0.254 -- 0.436 & 0.763 -- 0.986 \\
			$+$ Label Smoothing & 49.19 & 0.315 -- 0.463 & 0.635 -- 0.948 \\
			\bottomrule
		\end{tabular}
	\end{table}
	
\begin{figure}[t]
		\centering
		\includegraphics[width=\linewidth]{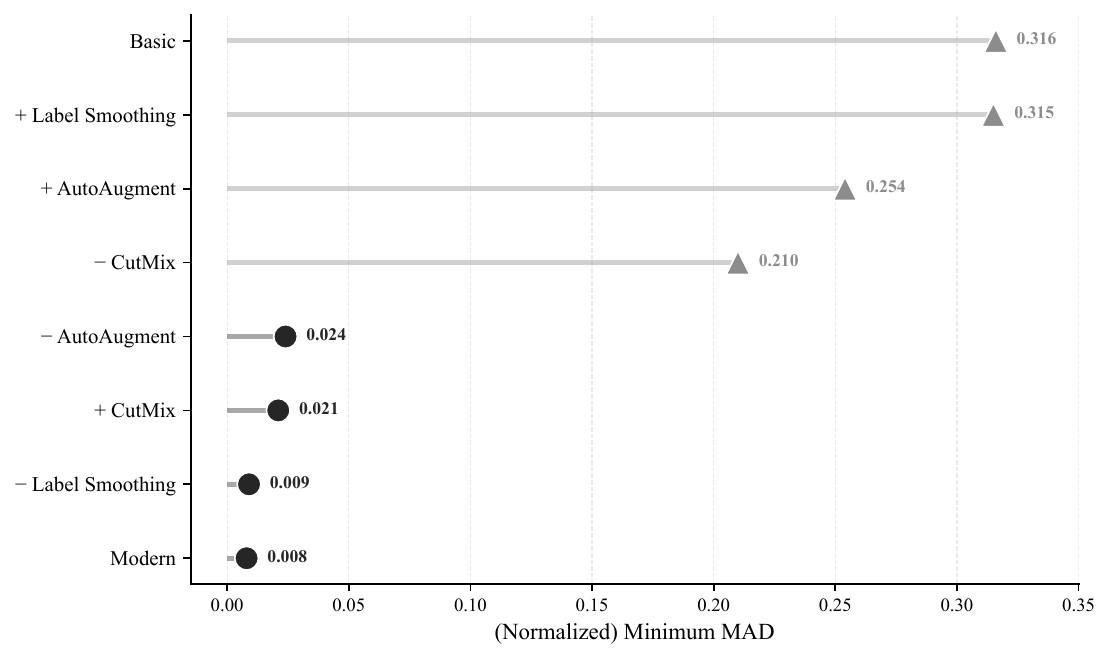}
		\caption{Minimum MAD per ablation condition, sorted in ascending order. Dark circles indicate conditions with CutMix and light triangles indicate conditions without CutMix. The binary separation demonstrates that CutMix is the determining factor for spatial locality in the conditions evaluated.}
		\label{fig:ablation_dotchart}
	\end{figure}

	The results reveal a clear pattern: CutMix is, within the experimental conditions evaluated, the only component that induces spatial locality. Figure~\ref{fig:ablation_dotchart} visualizes this finding by plotting the minimum MAD for each condition, differentiated by the presence of CutMix. The separation is striking: all conditions with CutMix (circles) achieve MAD $\leq 0.024$, while all conditions without CutMix (triangles) remain at MAD $\geq 0.210$. This represents nearly an order-of-magnitude separation between the two groups, with no overlap.
	
	To examine the effect in both directions, we consider removal and addition of each component. Since the design covers all possible combinations and all other hyperparameters remain fixed, the observed differences can be directly attributed to the manipulated component.
	
	Removing CutMix from the Modern protocol ($-$CutMix) increases the minimum MAD from 0.008 to 0.210, a $26\times$ increase. The locality pattern disappears despite retaining AutoAugment and Label Smoothing. Conversely, adding CutMix alone to the Baseline protocol ($+$CutMix) reduces the minimum MAD from 0.316 to 0.021, a $15\times$ reduction. Locality emerges without any other modern technique.
	
	In contrast, AutoAugment and Label Smoothing show no independent effect on locality. Adding only AutoAugment ($+$AutoAugment) produces minimum MAD $= 0.254$, nearly identical to the Baseline. Adding only Label Smoothing ($+$Label Smoothing) produces minimum MAD $= 0.315$, with virtually no change from the Baseline.
	
\begin{figure}[t]
		\centering
		\includegraphics[width=\linewidth]{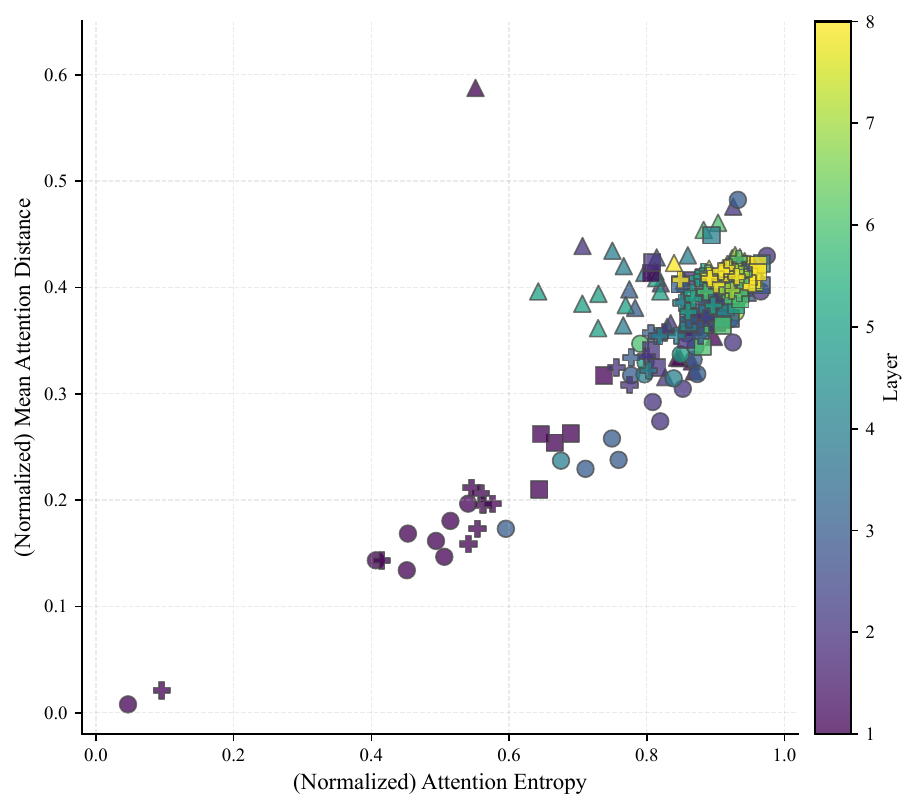}
		\caption{MAD vs.\ entropy for four key ablation conditions (CIFAR-100). Each marker represents an attention head: triangles for Baseline, circles for Modern, squares for $-$CutMix, and crosses for $+$CutMix. Color indicates layer depth (1--8). The $+$CutMix condition closely approximates the spatial pattern of the Modern protocol, while $-$CutMix collapses toward the Baseline.}
		\label{fig:ablation_scatter}
	\end{figure}

	Figure~\ref{fig:ablation_scatter} provides structural confirmation by plotting all attention heads in the MAD--entropy space for the four key conditions. The $+$CutMix condition (crosses) clusters with Modern (circles) in the low-MAD, low-entropy region characteristic of local and concentrated attention. Conversely, $-$CutMix (squares) collapses toward Baseline (triangles) in the high-MAD, high-entropy region. This demonstrates that the effect of CutMix is not limited to a single metric but transforms the attention structure across all layers.
	
	A notable dissociation emerges between classification accuracy and attentional locality. Both $+$CutMix and $+$AutoAugment improve test accuracy by comparable magnitudes over the Baseline (55.12\% and 55.22\%, respectively, vs.\ 48.53\%), yet only CutMix induces locality. This suggests that CutMix and AutoAugment improve performance through different mechanisms: CutMix by promoting the extraction of local features, AutoAugment by increasing input diversity without altering the geometry of attention.
	
	\subsection{Underlying Mechanism}
	
	The effect of CutMix on locality admits a direct interpretation. During training, CutMix replaces rectangular regions of one image with patches from another, forcing the model to classify correctly using only portions of each image. This pressure forces early layers to extract discriminative information from local neighborhoods, since global information may be corrupted or absent. As a result, attention heads specialized in local patterns emerge, precisely the inductive bias that characterizes convolutional networks. This result is consistent with the finding of Yun et al.~\cite{yun2019cutmix}, who reported that CutMix improves spatial localization ability in CNNs. Our work extends this finding to the context of Vision Transformers trained from scratch.
	
	In summary, the ablation study identifies CutMix as the only component that induces attentional locality in early layers. The separation between conditions with and without CutMix is binary (MAD $\leq 0.024$ vs.\ MAD $\geq 0.210$), with no overlap between them. AutoAugment and Label Smoothing do not alter this effect. This finding allows us to answer our initial question (whether the training protocol can induce locality) in a specific way: the need to classify from partial image regions is, within our experiments, the mechanism responsible.
	
	% ==================== 7. CONCLUSION ====================
	\section{Conclusion}
	\label{sec:concl}
	
	In this work we investigated whether the training protocol can induce spatial locality in Vision Transformers trained from scratch on moderate-scale datasets (50{,}000--100{,}000 images), without large-scale pretraining. Through a controlled comparison between Baseline and Modern protocols on three datasets (CIFAR-10, CIFAR-100, and Tiny-ImageNet), followed by a systematic ablation study on CIFAR-100, we identified a specific finding: CutMix is, within our experiments, the component responsible for the emergence of attentional locality in early layers.
	
	In our comparison (Section~\ref{sec:results}), we found that the Modern protocol is consistently associated with lower MAD and lower entropy in early layers, with the most pronounced effect on CIFAR-100 (minimum MAD of 0.008 versus 0.316 for the Baseline). The ablation study (Section~\ref{sec:ablation}) isolated the cause: CutMix is, within our experiments, the only component that produces this effect, with a binary separation and no overlap between conditions with CutMix (MAD $\leq 0.024$) and without CutMix (MAD $\geq 0.210$). AutoAugment and Label Smoothing showed no independent effect.
	
	CutMix forces the model to classify correctly from partial image regions, thereby pushing early layers to extract discriminative information from local neighborhoods. Yun et al.~\cite{yun2019cutmix} reported an analogous effect in CNNs, where CutMix improves spatial localization ability; our results suggest that this same principle operates in Vision Transformers, manifesting as the emergence of local attention in early layers.
	
	\paragraph{Implications.}
	From a practical standpoint, under the conditions evaluated CutMix was sufficient to induce attentional locality in early layers, without the need to combine multiple regularization techniques. From a theoretical standpoint, these findings complement the work of Raghu et al.~\cite{raghu2021dovit}, who established that data scale is a determining factor for the emergence of locality in ViTs. Our results suggest that the pressure to classify from partial regions constitutes an alternative factor capable of inducing local attention patterns on moderate-scale datasets.
	
	The observed independence between accuracy and locality is particularly informative. AutoAugment and CutMix improve test accuracy by comparable magnitudes ($\approx$6--7 percentage points over the Baseline), yet only CutMix alters the geometry of attention. This confirms that MAD and entropy capture structural properties genuinely distinct from classification performance, and that augmentation techniques with similar effects on accuracy can operate through fundamentally different internal mechanisms.
	
	\paragraph{Limitations and future work.}
	We acknowledge the following limitations. First, our experiments use a single compact architecture ($L=8$, $D=192$, $\approx$2.4M parameters); generalization to larger-scale models (e.g., ViT-B, ViT-L) requires independent validation. Second, all models are trained with a single initialization seed; although the metrics are computed over 5{,}000 images (which reduces estimation variance), the variability across independent training runs has not been quantified. Third, the ablation study was conducted exclusively on CIFAR-100; replication on CIFAR-10 and Tiny-ImageNet would strengthen the generality of the conclusion. Fourth, we did not analyze the temporal dynamics of locality emergence during training: at which epoch the local pattern appears and when it stabilizes.
	
	As future work, we propose the following directions: (i) multi-seed validation to quantify the variability of attention metrics across independent training runs; (ii) replication of the ablation study on the remaining datasets; (iii) extension to larger-scale ViT architectures; (iv) temporal analysis of locality emergence via intermediate checkpoints; and (v) evaluation of other image mixing strategies (e.g., Mixup, CutOut) to determine whether the causal effect resides specifically in the spatial nature of CutMix's rectangular cropping or in the more general principle of image mixing.
	
	% ==================== BIBLIOGRAPHY ====================
	\bibliographystyle{abbrvnat}
	\bibliography{references}
	
\end{document}